\documentclass[10pt,twocolumn,letterpaper]{article}

\usepackage{cvpr}              

\usepackage[font=normalfont,labelfont=bf]{caption}
\usepackage{algorithm}
\usepackage[accsupp]{axessibility}
\usepackage{algpseudocode}
\usepackage{color}
\usepackage[dvipsnames]{xcolor}
\usepackage{amsfonts} 
\usepackage{amsmath}
\usepackage{graphicx}
\usepackage{amssymb}
\usepackage{booktabs}
\usepackage{rotating}
\usepackage{multirow}
\usepackage{colortbl}
\usepackage{tabulary}
\usepackage{etoolbox}
\usepackage{pifont}
\usepackage{makecell}

\definecolor{LightOrange}{rgb}{1.0, 0.84, 0.4}
\usepackage{mathtools}
\usepackage{ragged2e}
\usepackage[pagebackref,breaklinks,colorlinks]{hyperref}

\usepackage[capitalize]{cleveref}
\crefname{section}{Sec.}{Secs.}
\Crefname{section}{Section}{Sections}
\Crefname{table}{Table}{Tables}
\crefname{table}{Tab.}{Tabs.}

\def\cvprPaperID{3} 
\def\confName{EarthVision}
\def\confYear{2023}

\begin{document}

\title{APPLeNet: Visual Attention Parameterized Prompt Learning for Few-Shot Remote Sensing Image Generalization using CLIP}

\author{Mainak Singha$^{1}\thanks{equal contribution}$ \and Ankit Jha$^{1*}$ \and Bhupendra Solanki$^{1}$ \and Shirsha Bose$^{2}$ \and Biplab Banerjee$^{1}$
\and
$^{1}$Indian Institute of Technology Bombay, India\and
$^{2}$Technical University of Munich, Germany
\and
{\tt\small \{mainaksingha.iitb, ankitjha16, bssiitb, shirshabosecs, getbiplab\}@gmail.com}}
\maketitle

\begin{abstract}
  In recent years, the success of large-scale vision-language models (VLMs) such as CLIP has led to their increased usage in various computer vision tasks. These models enable zero-shot inference through carefully crafted instructional text prompts without task-specific supervision. However, the potential of VLMs for generalization tasks in remote sensing (RS) has not been fully realized.
To address this research gap, we propose a novel image-conditioned prompt learning strategy called the Visual Attention Parameterized Prompts Learning Network (APPLeNet). APPLeNet emphasizes the importance of multi-scale feature learning in RS scene classification and disentangles visual style and content primitives for domain generalization tasks.
To achieve this, APPLeNet combines visual content features obtained from different layers of the vision encoder and style properties obtained from feature statistics of domain-specific batches. An attention-driven injection module is further introduced to generate visual tokens from this information. We also introduce an anti-correlation regularizer to ensure discrimination among the token embeddings, as this visual information is combined with the textual tokens.
To validate APPLeNet, we curated four available RS benchmarks and introduced experimental protocols and datasets for three domain generalization tasks. Our results consistently outperform the relevant literature and code is available at \url{https://github.com/mainaksingha01/APPLeNet}
\end{abstract}

\begin{figure}
    \centering
    \includegraphics[width=\columnwidth]{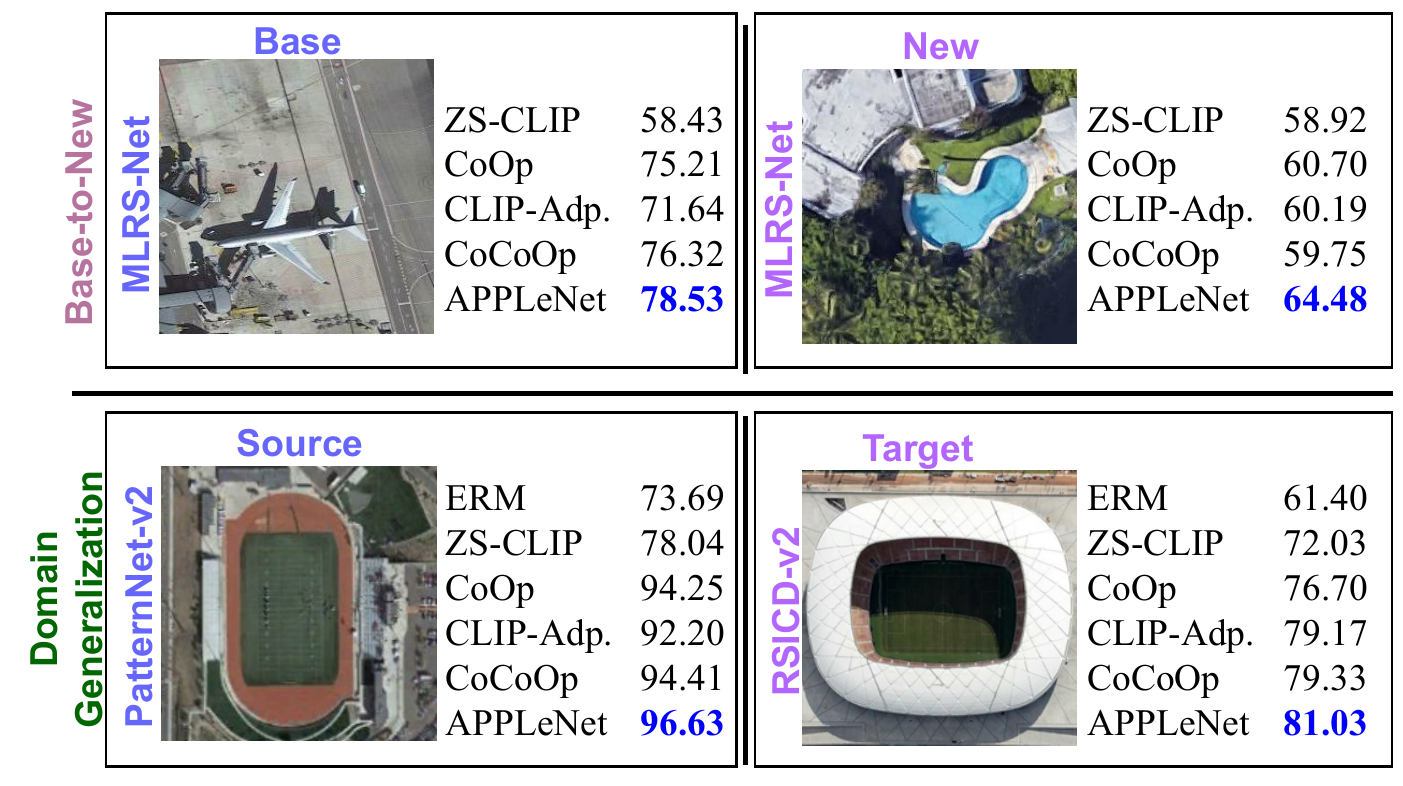}
    \vspace*{-8mm}
    \caption{\small{Performance overview of the proposed APPLeNet compared to state-of-the-art CLIP-based methods. It is shown that APPLeNet has better generalization capability irrespective of the complexity and size of the datasets in base-to-new class and across-domain generalization tasks.}}
    \label{applenet_teaser}
    \vspace*{-6mm}
\end{figure}

\section{Introduction}
\label{sec:intro}

Remote Sensing (RS) images play a vital role in numerous applications, including those mentioned in \cite{sabins1999remote, zhang2019joint, wu2017kernel, hino2018machine}. Traditional deep learning models have proven effective in recognizing complex RS images, outperforming ad-hoc machine learning techniques. However, these models tend to perform poorly in terms of generalization when faced with domain shifts. For example, Fig. \ref{applenet_teaser} illustrates this issue, where a model (ERM \cite{erm}) trained on images from the PatternNet \cite{li2018patternnet} dataset exhibits sub-optimal performance when applied to images from the RSICD \cite{lu2017exploring} dataset, captured by two sensors with divergent spatial characteristics.

To combat such changes in data distributions between training (source) and test (target) domains, researchers have investigated domain generalization (DG) \cite{li2018learning, rsdg1, zhou2022domain} and domain adaptation (DA) \cite{ganin2015unsupervised, tuia2016domain, farahani2021brief, saha2022multitarget, dars1, dars2}. DA follows a transductive setup, where the source and target domains are available simultaneously during training, while DG deals with a more realistic scenario, where a model trained on the source domain is applied to novel target domains during inference. Despite its success in computer vision literature, DG has yet to be thoroughly explored in RS.

From another perspective, few-shot learning (FSL) methods \cite{russwurm2020meta,zhang2021few, ji2022few, maml, chen2019meta} have emerged as a beneficial solution for alleviating the deep learning models' abundant data dependency for visual recognition, including in RS. FSL for different modalities, including multi-spectral and hyper-spectral, has been introduced in RS \cite{s3net,FSL_RS1,liu2018deep,spn}. However, these models are developed solely on image data and are suboptimal regarding the semantic richness of the embedding space learned by the feature extractors. As reported by these works, this significantly affects the FSL performance, and zero-shot transfer to novel tasks is not possible by FSL.

Large-scale pre-trained vision-language models (VLMs), such as CLIP \cite{clip}, ALIGN \cite{pmlr-v139-jia21b}, Florence \cite{yuan2021florence}, LiT \cite{zhai2022lit}, or Foundation models \cite{bommasani2021opportunities}, have recently shown promising results in generalizing to various downstream target domain tasks in a zero-shot manner with minimal supervision from a different source domain. These models align image-text pairs in a shared embedding space using a contrastive learning approach, making prompt engineering a critical aspect of VLMs.
However, manual prompt engineering is non-trivial, and prompt learning has received much attention to adapt CLIP for a target task. To address the generalization deficiency of the baseline prompt learning technique CoOp \cite{coop} and subsequent studies \cite{cocoop, prograd, clip-adapter} proposed to supplement the textual prompt embeddings with visual information extracted from CLIP's frozen vision encoder.
However, while these models are validated on natural image classification, we aim to explore their potential for scene recognition from optical RS images. This task is highly challenging due to the divergent spectral and spatial artifacts that characterize these images.

Although pre-trained CLIP \cite{clip} is highly effective, it falls short when evaluated on different domains, such as RS, as evidenced by the zero-shot CLIP's performance in Fig. \ref{applenet_teaser}. While prompt learning approaches like \cite{coop, cocoop, clip-adapter} improve on the performance of baseline CLIP, they are sub-optimal for cross-domain and cross-dataset generalization.
These approaches have three critical issues: \textbf{i}) they only consider visual features from the deepest layer and combine them with prompt token embeddings, ignoring low and mid-level features essential for optical RS scene classification where object scales are small, and texture plays a significant role in the classification task, \textbf{ii}) they add the same visual information to all token embeddings, causing redundancy, and \textbf{iii}) existing approaches fail to disentangle domain features from content features, which is likely to aid in DG.

Drawing from these discussions, this paper seeks to address two critical research questions: \textbf{i}) \textit{How can we effectively utilize CLIP's vision backbone to extract multi-scale visual content and style information for RS scenes to learn credible prompt tokens?} and \textbf{ii}) \textit{How can we guarantee that the learned prompt tokens contain non-redundant information?} We believe that tackling these issues together would result in more comprehensive and versatile prompts for optical RS scenes, as demonstrated in Fig. \ref{applenet_teaser}.

\noindent \textbf{Our proposed APPLeNet}: To address these challenges, our proposed approach, APPLeNet, makes three key contributions. Firstly, we utilize the intermediate blocks of CLIP's vision encoder to extract multi-scale visual content information. Secondly, we calculate the average feature representation for a batch of samples from a given domain to obtain style primitives for that domain. We leverage the concept of batch-norm statistics of feature embeddings from a CNN, which carry domain-specific knowledge \cite{instyle}. We combine the content and style features using an attention-based novel \textit{injection block} to generate dynamic image-conditioned visual tokens combining visual content and style properties. Finally, these tokens are added element-wise to the learnable text token embeddings to generate the prompts.

Thirdly, we introduce an anti-correlation regularizer to promote discrimination among prompt tokens. This regularizer penalizes high correlation among token embeddings. As a result, APPLeNet achieves more generalizable prompts than existing methods such as \cite{coop, cocoop, clip-adapter}. Notably, APPLeNet demonstrates strong performance even with extremely limited training data and is effective in different domain generalization scenarios with domain and label shifts.

 We highlight our \textbf{major contributions} as,

 \noindent [-] We propose a solution to the few-shot optical RS scene recognition and generalization problem by using pre-trained CLIP and introducing lightweight injection blocks in a model we call APPLeNet. The key innovations of APPLeNet are leveraging multi-scale visual content and style information from CLIP's vision encoder to learn prompt tokens and an anti-correlation regularizer that ensures the distinctiveness of the learned tokens.

\noindent [-] To validate our approach, we conduct extensive experiments on four optical RS image classification benchmarks and test for three essential and challenging generalization tasks: base-to-new class, cross-dataset, and single-source multi-target. We also introduce experimental protocols for these tasks, which have not been widely studied in RS.

Our experimental results demonstrate that APPLeNet outperforms the relevant literature substantially for all tasks by at least $2\%$ in mean classification scores.

\begin{figure*}
    \centering
    \includegraphics[scale=0.55]{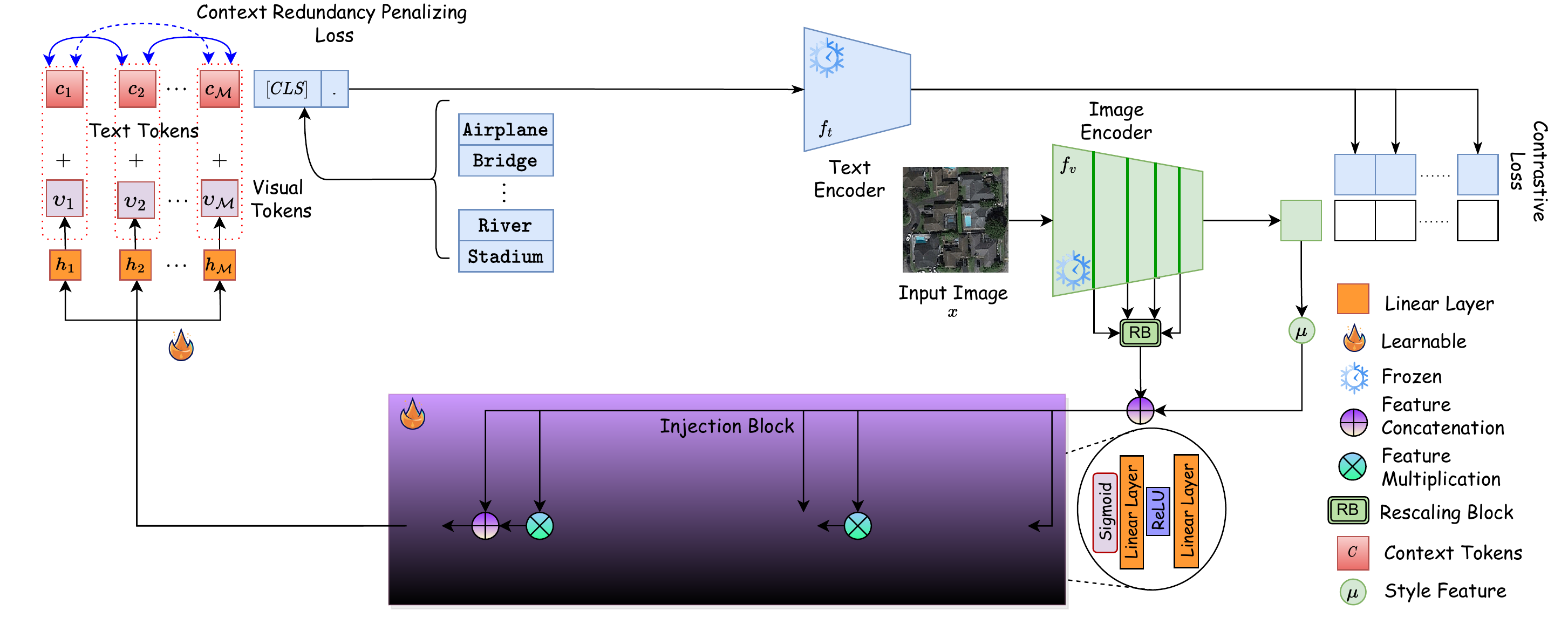}
    \vspace{-0.3cm}
    \caption{\small{APPLeNet is composed of a text encoder $f_{t}$, an image encoder $f_{v}$, and an injection block designed for multi-scale visual feature refinement. The $f_{v}$ produces multi-level visual content features, and the batch statistics $\mu$ for a domain as the style features, that are passed through a residual attention-based injection block. These features are then sent to individual projector networks $\{h_m\}_{m=1}^{\mathcal{M}}$ to derive the visual tokens $\{\upsilon_m\}_{m=1}^{\mathcal{M}}$. The visual tokens are added to the learnable text token embeddings $\{c_m\}_{m=1}^{\mathcal{M}}$ and, together with the class embeddings, are forwarded to $f_{t}$.
To reduce redundancy among the tokens, we introduce a novel Context Redundant Penalizing (CRP) loss ($\mathbf{L_{CRP}}$) among the context vectors in $\{c_m\}_{m=1}^{\mathcal{M}}$. The model is trained using a multi-task loss that comprises the contrastive loss between the image and prompt embeddings and the CRP regularizer.}}
    \label{applenet}
    \vspace{-0.4cm}
\end{figure*}


\section{Related Works}
\label{sec:related work}

\noindent \textbf{FSL in general and in RS:}
 Broadly speaking, existing few-shot learning (FSL) methods can be divided into transfer-learning based \cite{tlfsl1, tlfsl2}, meta-learning based \cite{hospedales2021meta, maml}, and metric-learning based \cite{snell2017prototypical, sung2018learning} approaches, respectively. Transfer learning fine-tunes the base model on each novel task but can underperform if the base and novel classes are from drastically different distributions. Alternatively, meta-learning-based supervised FSL approaches \cite{RS-MetaNet,ML_FSL} have gained attention because they can learn more generalizable features through episodic training. Metric-learning-based methods like matching networks \cite{vinyals2016matching}, prototypical networks \cite{snell2017prototypical}, and relation networks \cite{sung2018learning} focus on similarity optimization in a learnable fashion in episodes.

While all these approaches find their applicability in few-shot learning RS data, meta-learning-based algorithms \cite{FSL_RS1, RS-MetaNet} have been predominantly explored \cite{clip, ZENG2022143}. However, these models are prone to training class bias. One solution to tackle this is through uncertainty optimization, together with the FSL objective \cite{spn}.\\
\noindent \textbf{Domain generalization}: Deep learning models often face the challenge of domain shift between training and test distributions, which makes Domain Generalization (DG) a critical task for learning generalizable features from the training set that can be applied to any novel domain during inference. DG has three variants: single-source DG, multi-source DG, and heterogeneous DG \cite{l2a-ot, style-neophile, lostdg, sdg1, sdg2, hetdg1, hetdg2}. Initially, DG techniques focused on learning domain-invariant representations by considering data from multiple source domains through Domain Adaptation (DA) objectives \cite{li2020domain, ML_FSL, jia2020single, MMD}. Other approaches explored self-supervised learning \cite{jigen}, ensemble learning \cite{xu2014exploiting}, domain-specific networks \cite{mancini2018best}, and meta-learning \cite{learningtolearn}. However, training deep learning models for DG with limited samples may affect performance. Researchers have used augmentation methods \cite{sfa, l2a-ot}, and generative models \cite{style-neophile, mixstyle} to augment the source domains with diversified style primitives. However, DG for Remote Sensing (RS) image classification has received limited attention to date \cite{rsdg1, rsdg2}.

\textit{In contrast to existing FSL works in RS, which are trained from visual feature extractors, we are interested in leveraging the semantic superiority of CLIP-based foundation models for various target domain generalization tasks from a few source domain training samples}.

\noindent \textbf{Prompt learning for CLIP:}
Prompt learning is a widely used technique in natural language processing (NLP) \cite{petroni2019language}, which has recently made its way into the computer vision field. The main goal of prompt learning is to leverage pre-trained language models, such as BERT \cite{bert}, to provide valuable information for downstream tasks through prompts. Recent research has focused on automating the prompt generation process to eliminate manual interventions. One such approach is AutoPrompt \cite{shin2020autoprompt}, which explores tokens with the most significant gradient changes in the label likelihood. CoOp \cite{coop} optimizes prompts by fine-tuning CLIP for few-shot image classification. CoCoOp \cite{cocoop} proposes learning conditional prompts based on image features, partially improving CoOp's generalization capability.

In contrast, CLIP-Adapter \cite{clip-adapter} proposes fine-tuning feature adapters in both visual and language branches. ProGrad \cite{prograd} follows a similar approach to CoCoOp and explicitly ensures that the network remembers the knowledge learned from the foundation model. In \cite{tpt}, consistency among multiple views of the same image is used as supervision for prediction.

\textit{However, while \cite{cocoop, prograd} utilize visual information to improve prompts, they do not account for low to mid-level visual properties and visual style information in the prompt. Additionally, the learned tokens may contain redundant information. In contrast, our proposed APPLeNet addresses these issues and proposes a more comprehensive prompt learning strategy that is well-suited for handling RS scenes.}


\section{Proposed Methodology}
Let $\mathcal{D}_s = \{\mathcal{D}_s^i\}_{i=1}^{n}$ denote $n$ source domains, each with input data $x^i \in\mathcal{X}^i$ and corresponding label space $y^i\in\mathcal{Y}_{Seen}$. It is important to note that the probability distribution of each domain, $P(\mathcal{D}_s^i)$, may differ for all $i\in{1,\cdots,n}$. During training, we use the labels $\mathcal{Y}_{Seen}$ from $\mathcal{D}_s$, while during testing, we use $\mathcal{Y}_{Unseen}$ from a target test domain $\mathcal{D}_t$ with $P(\mathcal{D}_t) \neq P(\mathcal{D}_s^i)$, $\forall i\in{1,\cdots,n}$. For base-to-new class generalization, we set $\mathcal{Y}_{Seen}\cap\mathcal{Y}_{Unseen}=\emptyset$. In contrast, for domain generalization (DG), we consider single-source DG and assume that the label sets for both domains are identical ($\mathcal{Y}_{Seen}\cap\mathcal{Y}_{Unseen} = \mathcal{Y}_{Seen}\cup\mathcal{Y}_{Unseen}$).

Before presenting our proposed APPLeNet, we briefly introduce some important baselines, such as CLIP \cite{clip}, CoOp \cite{coop}, and CoCoOp \cite{cocoop}. 

\subsection{Relevant baselines}
\label{sec:baseline}
\noindent \textbf{CLIP:} CLIP \cite{clip} is a remarkable foundational model that learns an embedding space by seamlessly integrating visual and semantic knowledge. The model comprises two encoder heads: a visual encoder $f_v$ (either ResNet \cite{he2016deep}, or ViT \cite{vit}) for processing input images $x$, and a text encoder $f_t$ (BERT \cite{bert}) that considers the corresponding text prompt $t_y$ structured as \texttt{"a photo of $[\textit{CLS}]_y$"} where $[\textit{CLS}]_y]$ denotes the word embeddings for the class $y$. By means of contrastive training on a dataset of 400 million image-text pairs, CLIP strives to maximize the similarity between the image and the correct class prompt embeddings.

\noindent\textbf{CoOp:} CoOp \cite{coop} offers a solution to the issue of prompt engineering by replacing the manually created prompts with those generated through learning. This is achieved by utilizing a set of $\mathcal{M}$ learnable context vectors ${c_{1}, c_{2},\cdots, c_{\mathcal{M}}}$ that have the same dimensionality as the word embeddings and optimizing them through back-propagation. It's important to note that $\mathcal{M}$ is a hyperparameter that determines the context length and may differ between tasks. For any given class $y$, the prompt can be represented as $t_{y} = \{[c_{1}], [c_{2}],\cdots , [c_{\mathcal{M}}], [CLS_{y}]\}$.


\noindent\textbf{CoCoOp:} Despite the effectiveness of prompt learning, CoOp is susceptible to the domain-shift problem. CoCoOp \cite{cocoop} conditions prompt learning on visual features to mitigate this issue. This is achieved by introducing a meta-network that generates $\mathcal{M}$ meta-tokens, denoted as $\pi$. These meta-tokens are concatenated with context vectors to create prompts $t_{y} = \{[c_{1}(x)], [c_{2}(x)], \cdots , [c_{\mathcal{M}}(x)], [CLS_{y}]\}$, where $c_m(x) = c_{m}+\pi(x)$, $c_m$ is the $m^{th}$ text token. During CoCoOp's training, both the meta-network and context vector parameters are updated simultaneously.

\noindent \texttt{Important insight:} There have been various prompt learning techniques developed after CoOp and CoCoOp, such as \cite{prograd, maple}. However, these methods neglect two crucial factors in generalization tasks: the utilization of multi-scale feature composition from CLIP and the incorporation of visual style primitives into the prompts. These factors are particularly important in cases where there is a sudden change in style between source and target domains. The frozen image encoder ($f_v$) can be leveraged to encode style, while multi-scale content features can encode low, mid, and high-level visual properties, making them more transferable across categories.



\subsection{Our Proposed APPLeNet}
\label{sec:applenet}

Our paper introduces a new method called APPLeNet (Attention-Parameterized Prompt Learning Network) that leverages CLIP's visual backbone to extract multi-scale visual features and style features (mean $\mu$ of a batch of features from $f_v$) to improve text token learning.
APPLeNet, depicted in Figure \ref{applenet}, is composed of several critical components. First, it includes CLIP's frozen vision ($f_v$) and text ($f_t$) encoders. Additionally, it features a novel trainable \textit{Injection Block} (IB), indicated as $\mathcal{B}_{\phi}$, with learnable parameters $\phi$. This block emphasizes concatenated embeddings of style and multi-level visual features. Moreover, APPLeNet comprises $\mathcal{M}$ learnable linear layers that generate $\mathcal{M}$ visual tokens $\{\upsilon_{m}\}_{m=1}^{\mathcal{M}}$ given the outputs from $\mathcal{B}$. These visual tokens are then combined with the corresponding text tokens $\{c_m\}_{m=1}^{\mathcal{M}}$ which is further appended with the class token $[CLS_{y}]$ to generate prompt $t_y$. In the following sections, we elaborate on each component of APPLeNet.

\noindent\textbf{Encoding style and multi-scale content features into prompts:} To incorporate the multi-scale visual features from $f_v$ into $\mathcal{B}$, we propose using global average pooling (GAP) to collapse the spatial dimensions of each channel. This produces $\hat{f}_v^{l}(x) \in \mathbb{R}^{C \times 1}$, where $f_v^l \in \mathbb{R}^{W \times H \times C}$ represents the output responses from the $l^{th}$ layer. Here, $(W, H)$ represents the spatial dimensions of the feature maps. Using this approach, we define $\hat{F}(x) = [\hat{f}_v^1(x); \cdots ;\hat{f}_v^L(x)]$ as the concatenated multi-scale features obtained from all the $L$ encoder layers of $f_v$, where $[ ; ]$ denotes feature concatenation.

Furthermore, the average feature statistics corresponding to the batch of features from a domain act as the indicator for the style primitives. In this regard, let $\mu_i = f_v(X^i)$ represent the style for the $i^{th}$ domain.

Together we produce $F(x) = [\hat{F}(x_i);\mu_i]$, which captures both multi-scale content and the style information.

\noindent\textbf{Injection block:} The attention modules within $\mathcal{B}$ are denoted by ${\mathcal{A}_{q}(\cdot)}$, where $q \in {1,\cdots, \mathcal{Q}}$. For the first attention block ($q=1$), we denote the attended output features as $\mathcal{O}_1 = F(x) \odot \mathcal{A}_1 \oplus F(x)$. These features are then fed as input to $\mathcal{A}_2$ and so on, as follows:
\vspace{-0.22cm}
\begin{equation}
\mathcal{O}_{{q}} = 
\begin{dcases}
[ F(x)\odot\mathcal{A}_{q}(F(x))+F(x)],  \text{if } q = 1\\
[ \mathcal{O}_{q-1}\odot\mathcal{A}_{q}(\mathcal{O}_{q-1})+\mathcal{O}_{q-1}],  \text{otherwise}
\end{dcases}
\end{equation}
We subsequently pass $\mathcal{O}_{\mathcal{Q}}$ through $\mathcal{M}$ light-weight projector networks $\{h_m\}_{m=1}^{\mathcal{M}}$ which generate $\mathcal{M}$ visual tokens $\{\upsilon_1, \cdots,\upsilon_{\mathcal{M}} \}$: $\upsilon_m = h_m(\mathcal{O}_{\mathcal{Q}})$. We add the $m^{th}$ visual token embedding with the $m^{th}$ textual token embedding $c_m$: to obtain the $m^{th}$ prompt token embedding $c'_m = c_m + \upsilon_m$. The generated prompt is represented as:
\begin{equation}
    \centering
    t_y = \{[\upsilon_1 + c_1], \cdots, [\upsilon_{\mathcal{M}} + c_{\mathcal{M}}], [CLS_y]\}
\end{equation}

\subsection{Training and Inference}
We adopt a multi-task approach to train APPLeNet using two loss functions. The first one is the supervised contrastive loss, denoted as $\mathbf{L_{ce}}$, which ensures proper mapping between the visual feature representation $f_v(x)$ and the textual feature representation $f_t(t_y)$. This loss is formulated based on the cross-entropy approach. 

In addition, we introduce a context redundancy penalizing loss, denoted as $\mathbf{L_{CRP}}$. This loss ensures that the token embeddings in the set ${c_1 +v_1, \cdots, c_{\mathcal{M}}+ v_{\mathcal{M}}}$ do not carry redundant information. This helps the model learn a diverse set of tokens. In this regard, the prediction probability for $x$ to belong to the label $y$ is denoted by,
\begin{equation}
p(y|x) = \frac{ \exp(\text{sim}(f_v(x), f_t(t_y(\mathcal{B}_{\phi}(x)))/\tau))}{\sum_{k=1}^{|\mathcal{Y}|}\ \exp(\text{sim}(f_v(x), f_t(t_k(\mathcal{B}_{\phi}(x)))/ \tau))}
\end{equation}

`$\text{sim}$' denotes the \texttt{cosine} similarity, and $\tau$ is the temperature hyper-parameter.
The cross-entropy loss ($\mathbf{L_{ce}}$) is computed between the prediction probabilities of each input image and their corresponding class labels as follows:
\begin{equation}
    \mathbf{L_{ce}} = \underset{\mathcal{B}_{\phi}, \{h_m\}}{\arg\min} \underset{(x,y) \in \mathcal{P}(\mathcal{D}_s)}{\mathbb{E}}  - \sum_{k=1}^{\mathcal{Y}_{Seen}} y_{k} log(p(y_k|x))
\end{equation}

Simultaneously, we seek to decorrelate pairwise the token embeddings using $\mathbf{L_{CRP}}$ as,
\begin{equation}
\begin{aligned}
\mathbf{L_{CRP}} & = \underset{\mathcal{B}_{\phi},\{ h_m\}}{\arg\min}  \underset{(x,y) \in \mathcal{P}(\mathcal{D}_s)}{\mathbb{E}} \begin{vmatrix}c'_{j}(x)\cdot c'_{l}(x) - \mathcal{I}\end{vmatrix}, \\ & \forall j,l \in \{1,2,\cdots,\mathcal{M}\}, j \neq l, c'_j =c_j + v_j
\end{aligned}
\end{equation}

Hence, the total loss ($\mathbf{L_{total}}$) is computed as:
\begin{equation}
    \mathbf{L_{total}} = \underset{\mathcal{B}_{\phi}, \{h_m\}}{\arg\min} [\mathbf{L_{ce}} + \lambda * \mathbf{L_{CRP}}]
    \label{eq_7}
\end{equation}
Where $\lambda$ is the weighting hyper-parameter. In the inference stage, we compute the cosine similarity between the images $x_t \in \mathcal{D}_t$ and prompt embeddings for all the classes in $\mathcal{Y}_{Unseen}$. The class with a high probability value is selected.
\begin{equation}
    \hat{y_t} = \underset{y \in \mathcal{Y}_{Unseen}} {\arg \max} p(y|x_t)
\end{equation}

\section{Experimental evaluations}

\noindent \textbf{Dataset descriptions:} Our experimental evaluation involves four datasets: PatternNet \cite{li2018patternnet}, RSICD \cite{lu2017exploring}, RESISC45 \cite{cheng2017remote}, and MLRSNet \cite{qi2020mlrsnet}.

PatternNet comprises 38 classes, with each class containing 800 images of size 256 $\times$ 256 pixels. RSICD includes 30 classes and a total of 10,000 images, each with a size of 224 $\times$ 224 pixels. Notably, each class has a different number of images.

RESISC45 consists of 45 classes, with each class containing 700 images of size 256 $\times$ 256 pixels. MLRSNet comprises 46 classes and a total of 109,161 images, each with a size of 256 $\times$ 256 pixels.

Furthermore, we extend our work to generate learnable prompts in the single-source multi-target domain generalization setup. In this regard, we curate new versions (v2) of the above-mentioned datasets, where we consider the 16 overlapping classes from all four datasets. Details are mentioned in the \texttt{supplementary text}.

\noindent \textbf{Architecture Details:}
In all our experiments, $\mathcal{B}_{\phi}$ comprises two attention modules, each followed by a linear layer. Our attention module is inspired by SE-Net \cite{hu2018squeeze} and has two linear layers, each followed by \texttt{ReLU} and \texttt{Sigmoid} activation functions, respectively. However, we can accommodate more attention blocks in $\mathcal{B}$, if required. Further, each $h_m$ is designed as a single dense layer, which converts $\mathcal{B}_{\phi}(x)$ into dimensions equal to the text embeddings.

\begin{table*}[ht!]
\centering
\scriptsize{
    \centering
    \caption{\small{Comparison of APPLeNet with state-of-the-art methods for base-to-new (B2N) class generalization task. We indicate the validation accuracy for the Base and New classes. H denotes the harmonic mean used to generalize the trade-off performance between the base and new classes. Best results are shown in \textbf{bold}.}}
    \vspace{-0.2cm}
    \scalebox{1.03}{
    \begin{tabular}{lccccccccccccccc} 
    \toprule
	
\rowcolor{gray!20}&\multicolumn{3}{c}{\textbf{PatternNet}}&\multicolumn{3}{c}{\textbf{RSICD}} &\multicolumn{3}{c}{\textbf{RESISC45}}
    &\multicolumn{3}{c}{\textbf{MLRSNet}}&\multicolumn{3}{c}{\textbf{Avg. of all}}\\
      \cmidrule(lr){2-4}\cmidrule(lr){5-7}\cmidrule(lr){8-10}\cmidrule(lr){11-13}\cmidrule(lr){14-16}
     
  \rowcolor{gray!20} \multirow{-2}{*}{\textbf{Method}} &\multicolumn{1}{c}{\textbf{Base}}&\multicolumn{1}{c}{\textbf{New}}&\multicolumn{1}{c}{\textbf{H}}
    &\multicolumn{1}{c}{\textbf{Base}}&\multicolumn{1}{c}{\textbf{New}}&\multicolumn{1}{c}{\textbf{H}}
    &\multicolumn{1}{c}{\textbf{Base}}&\multicolumn{1}{c}{\textbf{New}}&\multicolumn{1}{c}{\textbf{H}}
    &\multicolumn{1}{c}{\textbf{Base}}&\multicolumn{1}{c}{\textbf{New}}&\multicolumn{1}{c}{\textbf{H}}
    &\multicolumn{1}{c}{\textbf{Base}}&\multicolumn{1}{c}{\textbf{New}}&\multicolumn{1}{c}{\textbf{H}}\\
    
    \midrule
    \cellcolor[gray]{0.9}CLIP \cite{clip} & 63.67 & 64.37 & 64.02 & 54.61 & 55.33 & 54.97 & 56.32 & 55.38 & 55.85 & 51.43 & 51.92 & 51.67 & 56.51 & 56.75 & 56.63 \\

    \cellcolor[gray]{0.9}CoOp \cite{coop}   & 91.62 & 62.23 & 74.12 & 92.52 & 56.08 & 69.83 & 89.04 & 55.75 & 68.57 & 75.21 & 53.64 & 62.62 & 87.10 & 56.93 & 68.85 \\

    \cellcolor[gray]{0.9}CLIP-Adapter \cite{clip-adapter} & 82.15 & 63.26 & 71.48 & 78.93 & 55.44 & 65.13 & 81.67 & 56.23 & 66.60 & 71.64 & 53.19 & 61.05 & 78.60 & 57.03 & 66.10 \\

    \cellcolor[gray]{0.9}CoCoOp \cite{cocoop}& 92.39 & 63.34 & 75.16 & 93.18 & 58.67 & 72.00 & 89.78 & 57.18 & 69.86 & 76.32 & 52.75 & 62.38 & 87.92 & 57.99 & 69.88 \\

    \cellcolor[gray]{0.9}ProGrad \cite{prograd} & 92.65 & 62.48 & 74.63 & 93.44 & 58.15 & 71.69 & 90.13 & 57.89 & 70.50 & 75.96 & 52.23 & 61.90 & 88.05 & 57.69 & 69.70 \\

    \cellcolor{red!25}APPLeNet & \textbf{94.89} & \textbf{65.57} & \textbf{77.55} & \textbf{95.26} & \textbf{60.71} & \textbf{74.16} & \textbf{91.24} & \textbf{60.46} & \textbf{72.73} & \textbf{78.53} & \textbf{56.41} & \textbf{65.66} & \textbf{89.98} & \textbf{60.79} & \textbf{72.56} \\ \bottomrule
    \end{tabular}}\label{B2N}}
    \vspace{-0.4cm}
\end{table*}
\noindent \textbf{Training and evaluation protocols:}
We train APPLeNet for $50$ epochs using the stochastic gradient descent (SGD) optimizer \cite{robbins1951stochastic} with an initial learning rate of $2e^{-4}$ and a warm-up fixed learning rate of $1e^{-7}$ during the first epoch to prevent explosive gradients. We keep ViT-B/16 as the image encoder backbone, use $16$ training samples (i.e., shots) from each class, and create a batch size of $4$ and $\lambda$ (Eq. \ref{eq_7}) to be $0.1$ for model training. We initialize the text prompts from the embeddings of \texttt{"a photo of a [CLS]"} which means the context length is four. This follows the previous literature \cite{coop, cocoop}. We execute the model using three seeds and report the average \texttt{top-1} accuracy.


\subsection{\textbf{Comparison with the state-of-the-art methods}}
In this section, we discuss the performance of APPLeNet with respect to the methods from the literature for the three DG tasks, as mentioned: i) \textbf{Base-to-new class generalization}, where the training and test classes are disjoint. ii) \textbf{Cross-dataset generalization}, where the model is trained on one dataset and evaluated on novel datasets with domain and label shifts. iii) \textbf{Single source multi-target DG}, where the model is trained on a source domain and evaluated on multiple novel domains under the closed-set setting.

\noindent \textbf{Baselines}: We evaluated the performance of APPLeNet compared to existing methods from the prompting literature using CLIP. As a baseline, we used Zero-shot CLIP \cite{clip}. In addition, we explored other approaches, such as ERM \cite{erm}, which involves a trainable linear model on top of CLIP. We also tested a state-of-the-art DA technique, DANN \cite{dann}, in combination with the CLIP features. Finally, we examined prompt learning techniques, including CoOp \cite{coop}, CoCoOp \cite{cocoop}, CLIP-Adapter \cite{clip-adapter}, and ProGrad \cite{prograd}.

\noindent\textbf{Base-to-New (B2N) class generalization:} Table \ref{B2N} presents the experimental results for B2N class generalization on the four RS datasets, where the harmonic mean (H) between the classification accuracies of the Base and New classes is computed. For all the datasets, we randomly and equally divide the datasets into two groups to define the source (with the base classes) and the target (with the novel classes) domains.
Compared to the CLIP's zero-shot approach, APPLeNet achieves better generalization scores, with a considerable margin of $33.47\%$ on seen classes and $4.04\%$ on unseen classes over all datasets (on average). We also compare APPLeNet with referred context optimization-based methods, where it outperforms CoOp and CoCoOp on the PatternNet \cite{li2018patternnet}, RSICD \cite{lu2017exploring}, RESISC45 \cite{cheng2017remote}, and MLRSNet \cite{qi2020mlrsnet} datasets by $3.4\%, 4.3\%, 4.2\%$, and $3.0\%$, and $2.4\%$, $2.2\%$, $2.9\%$, and $3.3\%$, respectively. In PatternNet, APPLeNet consistently beats CoCoOp by huge margins of $5.8\%, 5.2\%$ and $6.4\%$ in \textit{river}, \textit{storage tank} and \textit{tennis court} classes. Among all the referred methods, only CoCoOp and ProGrad show the second and third-best performance scores on generalizing the unseen classes over all RS datasets.

\noindent\textbf{Cross-Dataset (CD) generalization:} Table \ref{CDT} presents the results of our evaluation of APPLeNet on the CD setup. In this regard, we train the model on the PatternNet \cite{li2018patternnet} dataset (source domain) and report zero-shot inference results on the remaining RS datasets (target domains). Our APPLeNet outperforms the source and target classification performance by significant margins of $26.5\%$ and $13.9\%$, respectively, compared to zero-shot (CLIP) and non-learnable prompt (CLIP-Adapter) methods. Besides, APPLeNet outperforms CoCoOp by $1.3\%$, $1.4\%$, and $2.1\%$ for the unseen RS domains, namely RSICD \cite{lu2017exploring}, RESISC45 \cite{cheng2017remote}, and MLRSNet \cite{qi2020mlrsnet} datasets, respectively. Also APPLeNet beats CoCoOp by $3.6\%, 5.2\%, 5.4\%$ and $4.7\%$ in \textit{desert}, \textit{mountain}, \textit{port} and \textit{school} of RSICD dataset. Finally, AppLeNet is better than ProGrad than at least $2.5 \%$ on all the target tasks. Based on these results, our results establish that APPLeNet successfully narrows the generalization gap between a single source and multiple targets with domain and label shifts in the CD transfer technique.

\begin{table}[!ht]
\centering
\scriptsize{
    \centering
    \caption{\small{Comparison of APPLeNet with state-of-the-art methods for cross-dataset generalization with PatternNet dataset as the source domain and remaining RS datasets as the target domains. We use the accuracy metric as the performance measure. Best results are shown in \textbf{bold}.}}
    \vspace{-0.2cm}
    \scalebox{0.93}{
    \begin{tabular}{lcccc} 
    \toprule
   \rowcolor{gray!20} &\multicolumn{1}{c}{\textbf{Source}}&\multicolumn{3}{c}{\textbf{Target}} \\
     
    \cmidrule(lr){2-2}\cmidrule(lr){3-5}
     
  \rowcolor{gray!20}  \multirow{-2}{*}{\textbf{Method}}&\multicolumn{1}{c}{\textbf{PatternNet}}&\multicolumn{1}{c}{\textbf{RSICD}}&\multicolumn{1}{c}{\textbf{RESISC45}}
    &\multicolumn{1}{c}{\textbf{MLRSNet}}\\
    
    \midrule
    \cellcolor{gray!25}CLIP \cite{clip} & 61.72 & 43.25 & 48.56 & 45.13 \\

    \cellcolor{gray!25}CoOp \cite{coop} & 85.23 & 42.53  & 49.34 & 44.50  \\

    \cellcolor{gray!25}CLIP-Adapter \cite{clip-adapter} & 74.27& 42.57  & 49.07  & 44.17  \\

    \cellcolor{gray!25}CoCoOp \cite{cocoop} & 85.95 & 43.61 & 49.53 & 44.72\\

    \cellcolor{gray!25}ProGrad \cite{prograd} & 86.14 & 41.25 & 48.26 & 44.12 \\

   \cellcolor{red!25} APPLeNet & \textbf{88.17} & \textbf{44.87} & \textbf{50.97} & \textbf{46.83} \\ \bottomrule
    \end{tabular}}\label{CDT}}
    \vspace{-0.4cm}
\end{table}

\noindent\textbf{Single source multi-target domain generalization (DG):}
We tested the generalization performance of our proposed APPLeNet on a Single-Source Multi-Target (SSMT) DG setup. Unlike the CD setting discussed earlier, we only considered the common classes across all datasets since SSMT is a closed-set setting. We trained the model on the PatternNetv2 dataset and evaluated it on the remaining datasets. The comparison results with the state-of-the-art (SOTA) methods and APPLeNet are presented in Table \ref{DG}.

The results show that ProGrad outperformed other referenced prompting techniques by at least $0.6\%$, while APPLeNet surpassed all of them by a minimum margin of $2.4\%$ on the MLRSNetv2, $1.6 \%$ on RSICDv2, and $1.4 \%$ on RESISC45v2 (target domains), respectively. APPLeNet beats CoCoOp in \textit{beach}, \textit{forest}, and \textit{river} classes with a huge margin of $5.5\%, 4.8\%$ and $5.9\%$ on average over the target datasets. Notably, APPLeNet effectively transferred the learned classification information from the PatternNetv2 to classes such as \textit{desert}, \textit{sparse residential}, and \textit{river} to the RESISC45v2 and outperformed the SOTA methods by at least $5.7\%$.

Regarding the source domain classification task, APPLeNet achieved a performance of $88.17 \%$, which was better than the second-best by $2.03 \%$.
\begin{table}[!ht]
\centering
\scriptsize{
    \centering
    \caption{\small{Comparing APPLeNet with state-of-the-art methods for single-source multi-target domain generalization on our released ($2^{nd}$ version) benchmark RS datasets. We use the accuracy metric as the performance measure. Best results are shown in \textbf{bold}.}}
    \vspace{-0.2cm}
    \scalebox{0.91}{
    \begin{tabular}{lcccc} 
    \toprule
   \rowcolor{gray!20} &\multicolumn{1}{c}{\textbf{Source}}&\multicolumn{3}{c}{\textbf{Target}} \\
     
    \cmidrule(lr){2-2}\cmidrule(lr){3-5}
     
   \rowcolor{gray!20}   \multirow{-2}{*}{\textbf{Method}}&\multicolumn{1}{c}{\textbf{PatternNetv2}}&\multicolumn{1}{c}{\textbf{RSICDv2}}&\multicolumn{1}{c}{\textbf{RESISC45v2}}
    &\multicolumn{1}{c}{\textbf{MLRSNetv2}}\\
    
    \midrule
  \cellcolor[gray]{0.9}  ERM \cite{erm} & 73.69  & 61.40  & 61.59 & 61.13  \\
   \cellcolor[gray]{0.9} CLIP \cite{clip} & 78.04 & 72.15 & 75.42 & 67.78 \\
    
   \cellcolor[gray]{0.9} DANN \cite{dann} & 93.56 & 75.49  & 76.18  & 70.53  \\

   \cellcolor[gray]{0.9} CoOp \cite{coop} & 94.25 & 76.50  & 77.87& 70.97  \\

    \cellcolor[gray]{0.9}CLIP-Adapter \cite{clip-adapter} & 92.36 & 79.17 & 79.76 & 71.04  \\

 \cellcolor[gray]{0.9}   CoCoOp \cite{cocoop} & 94.41 & 79.33 & 80.43 & 71.67 \\

    \cellcolor[gray]{0.9}ProGrad \cite{prograd} & 95.18 & 77.46 & 80.65 & 72.29  \\

   \cellcolor{red!25} APPLeNet & \textbf{96.63} & \textbf{81.03} & \textbf{82.23} & \textbf{74.03} \\ \bottomrule
    \end{tabular}}\label{DG}}
    
\end{table}

\subsection{Ablation analysis}
\label{ablation}
\begin{figure}[ht!]
    \centering
    \vspace{-0.4cm}
    \includegraphics[scale=0.0182]{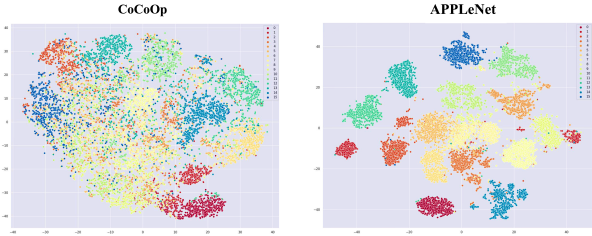}
    \vspace{-0.8cm}
    \caption{\small{t-SNE plots \cite{van2008visualizing} for the image feature extracted from the Meta-Net of CoCoOp and the Injection Block (IB) of APPLeNet for the SSMT domain generalization task on the MLRSNetv2 dataset. The legends represent the class labels.}}
    \label{tsne}
    \vspace{-0.3cm}
\end{figure}

\noindent\textbf{t-SNE visualization:} In Figure \ref{tsne}, we present a t-SNE \cite{tsne} visualization of the image embeddings generated by APPLeNet and compare them with CoCoOp \cite{cocoop} on the MLRSNetv2 dataset for the SSMT-DG task. The visualization clearly demonstrates that APPLeNet can accurately cluster each class, while the cluster points of many classes get overlapped in CoCoOp. This confirms the discriminability of APPLeNet.

\noindent\textbf{Sensitivity to the variation in the number of shots:} We evaluate the performance of our proposed APPLeNet by varying the number of shots from 1 to 32 for the B2N class generalization task and compare it with the state-of-the-art (SOTA) prompting techniques, as shown in Table \ref{shots}. In this setting, we use a context length of 4, place the class token at the end, use ViT-B/16 as the visual feature backbone, and use a unified context vector. As CLIP is a zero-shot approach, we exclude it and only consider few-shot-based prompting methods to compare and show results on the PatternNet dataset.

We are able to outperform the benchmark prompt learning-based methods by at least $0.8\%$, $2.4\%$, and $1.6\%$ for 8, 16, and 32 shots, respectively.

\begin{table}[!ht]
\centering
\scriptsize{
    \centering
    \caption{\small{Comparison of APPLeNet with state-of-the-art methods on varying the number of shots for the B2N class generalization task with PatternNet dataset. Harmonic mean (H) of base and new classes is considered for comparison, as well as to depict the generalization trade-off. Best results are shown in \textbf{bold}.}}
    \vspace{-0.2cm}
    \scalebox{1.0}{
    \begin{tabular}{lccccc} 
    \toprule
   \rowcolor{gray!20} \multirow{1}{*}{\textbf{Method}}&\multicolumn{1}{c}{\textbf{1-shot}}&\multicolumn{1}{c}{\textbf{4-shots}}
    &\multicolumn{1}{c}{\textbf{8-shots}}
    &\multicolumn{1}{c}{\textbf{16-shots}}
    &\multicolumn{1}{c}{\textbf{32-shots}} \\
     
     
    
    \midrule

    \cellcolor{gray!25}CoOp \cite{coop} & 70.33 & 71.61 & 72.17 & 74.12 & 74.58  \\

    \cellcolor{gray!25}CLIP-Adapter \cite{clip-adapter} & 69.75 & 69.95 & 70.37 & 71.48 & 71.64  \\

    \cellcolor{gray!25}CoCoOp \cite{cocoop} & 71.85 & \textbf{73.61} & 74.53 & 75.16 & 74.39 \\

    \cellcolor{gray!25}ProGrad \cite{prograd} & \textbf{73.67} & 72.05 & 73.16 & 74.63 & 75.56  \\

    \cellcolor{red!25}APPLeNet & 72.44 & 72.46 & \textbf{75.28} & \textbf{77.55} &
    \textbf{77.13}\\ \bottomrule
    \end{tabular}}\label{shots}}
    \vspace{-0.3cm}
\end{table}

\begin{figure}
    \centering
    \vspace{-0.2cm}
    \includegraphics[width=8.6cm]{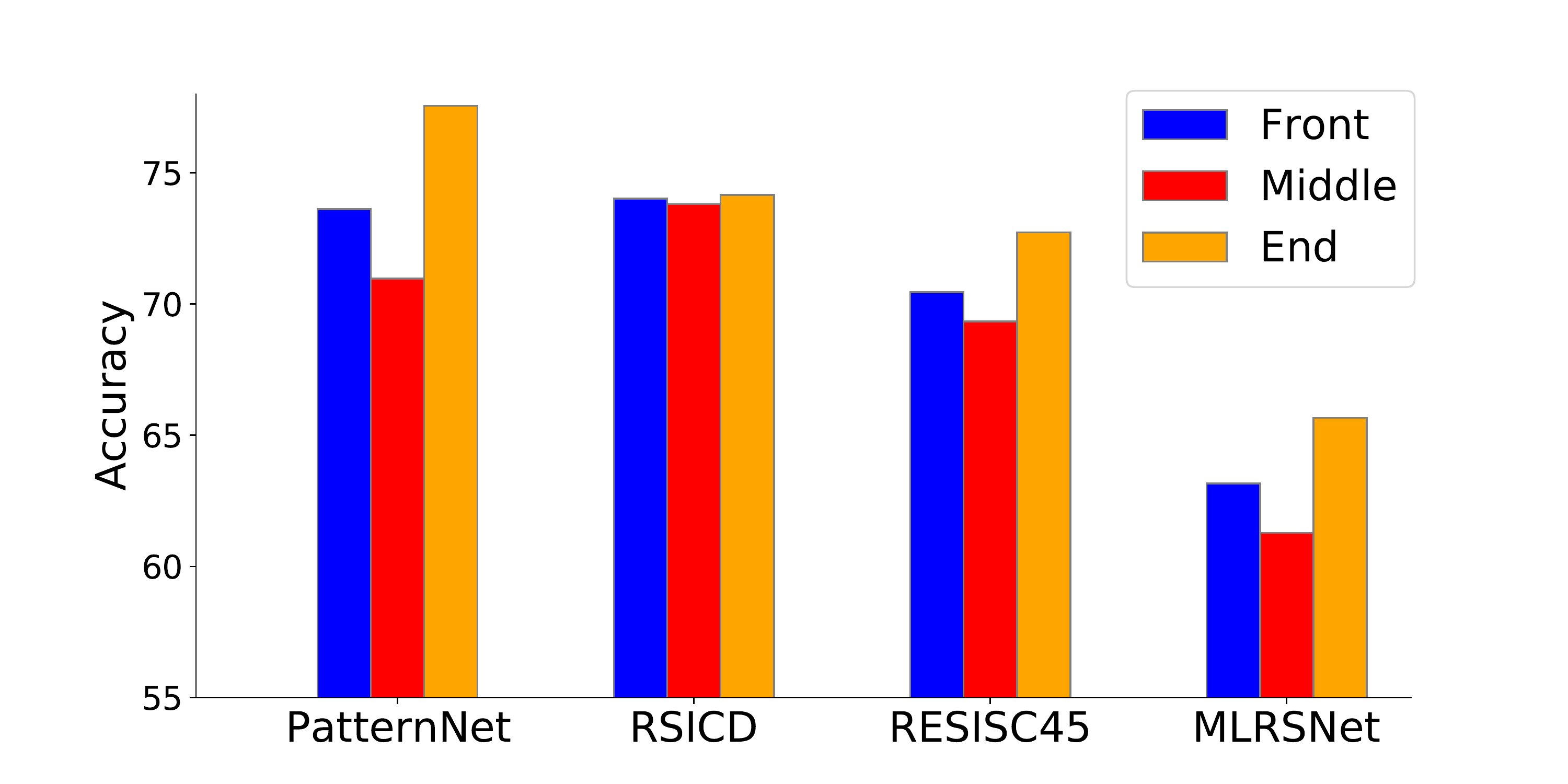}
    \vspace{-0.9cm}
    \caption{\small{Classification performance on changing position of the class tokens in APPLeNet, i.e., `Front',`Middle', and `End' for the B2N class generalization task on the four RS datasets. We consider the harmonic mean (H) of base and new classes for comparison.}}
    \label{fig:class_position}
    \vspace{-0.3cm}
\end{figure}
    
\begin{figure}
    \centering
    \includegraphics[width=8.8cm]{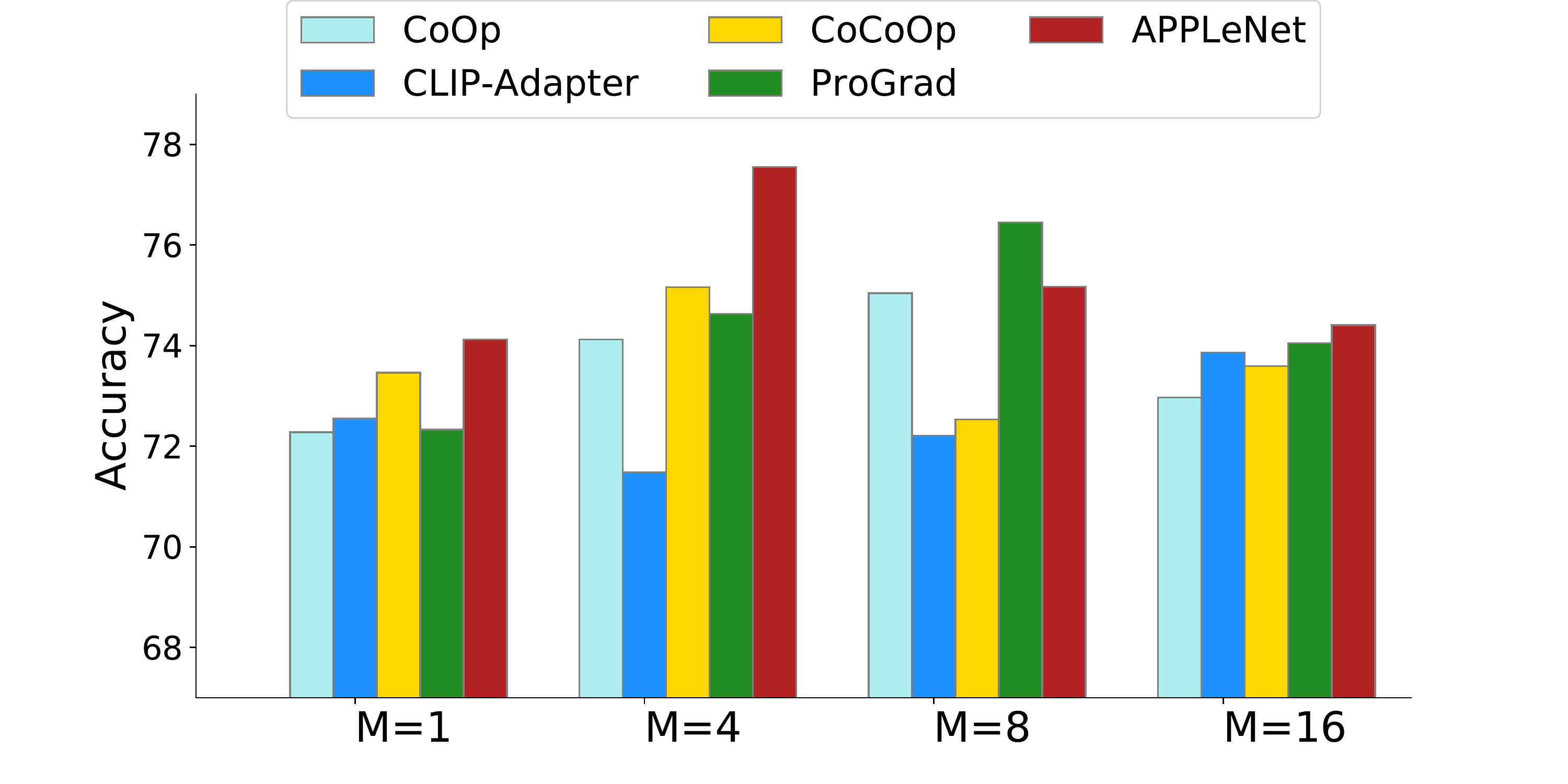}  \vspace{-0.8cm}
    \caption{\small{Classification performance of APPLeNet by varying the context length (M) for the B2N class generalization on PatternNet dataset and compared with the SOTA methods. The harmonic mean (H) of base and new classes are considered for comparison.}}
    \label{fig:context_length}
    \vspace{-0.4cm}
\end{figure}

\begin{figure}
    \centering
    \includegraphics[width=0.44\textwidth]{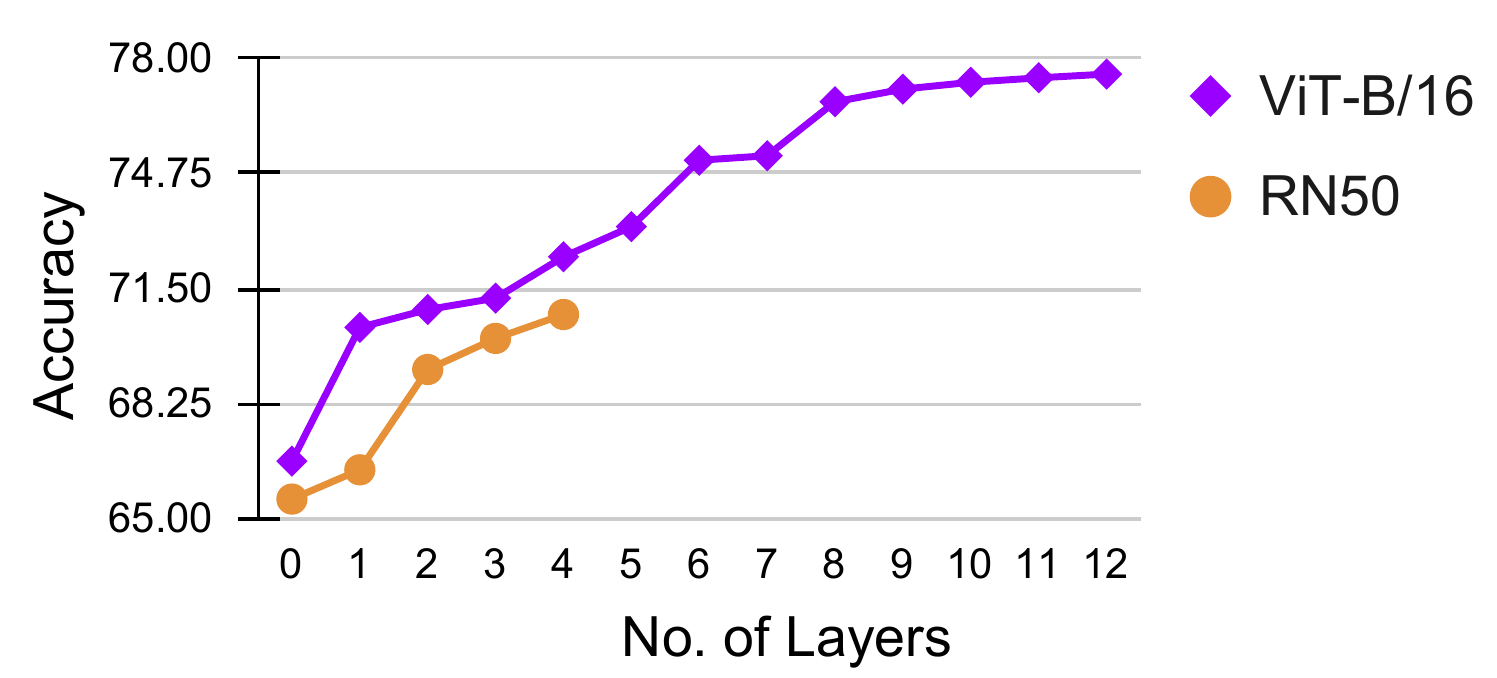}
    \vspace{-0.34cm}
    \caption{\small{Performance of APPLeNet with different layers of ViT-B/16 and RN50 backbones to extract multi-scale features on PatterNet dataset. Harmonic mean (H) of base and new classes are shown as accuracy.}}
    \label{fig:class_multi}
    \vspace{-0.4cm}
\end{figure}
\begin{table}[!ht]
\centering
\scriptsize{
    \centering
    \caption{\small{Ablation of different types of initialization of context vectors in APPLeNet. Harmonic mean (H) of base and new classes are considered for comparison. Best results are shown in \textbf{bold}.}}
    \vspace{-0.2cm}
    \scalebox{0.93}{
    \begin{tabular}{lcccc} 
    \toprule
    \multirow{1}{*}{\textbf{Context Vectors}}&\multicolumn{1}{c}{\textbf{PatternNet}}&\multicolumn{1}{c}{\textbf{RSICD}}
    &\multicolumn{1}{c}{\textbf{RESISC45}}
    &\multicolumn{1}{c}{\textbf{MLRSNet}} \\
     
     
    
    \midrule

    manual initialization & 77.55 & \textbf{74.16}  & \textbf{72.73} & \textbf{65.66} \\

    random initialization & \textbf{81.61} & 68.58 & 70.80 & 61.45 \\ 
    no initialization & 67.90 & 69.57 & 69.42 & 59.16 \\ \bottomrule
    \end{tabular}}\label{crploss2}}
    \vspace{-0.2cm}
\end{table}
\noindent\textbf{Sensitivity to the position of the class token and the prompt initialization strategy:}
In this experiment, we investigate the effect of the position of the class token in the learnable context vectors in $[1,\mathcal{M}]$ on the performance of APPLeNet in the B2N class generalization task. We experiment with three different positions for the class token: ``front", ``middle", and ``end", while generating the learnable prompts.

We plot the harmonic mean between the base and new classes for all the RS datasets in Figure \ref{fig:class_position}. Our results show that positioning the class token at the "end" of the context vector consistently improves the performance of APPLeNet on the B2N class generalization task, with at least a $3.9\%$, $2.3\%$, and $2.5\%$ improvement on PatternNet, RESISC45, and MLRSNet datasets, respectively, compared to positioning the token at the ``front" or ``middle". However, on the RSICD dataset, we observe no significant difference in performance for the different class token positions.

Finally, we consider three different prompt initialization strategy to check their efficacy in Table \ref{crploss2} for B2N Generalization. It highlights that manual initialization from \texttt{"a photo of a"} outperforms the random initialization and no initialization strategies significantly for the target datasets. Interestingly, the random initialization outperforms other on the source domain evaluation.

\noindent\textbf{Sensitivity analysis of APPLeNet to context lengths ($\mathcal{M}$):}
During test-time prompt generation, we varied the context length ($\mathcal{M}$) and experimented with four different context lengths: 1, 4, 8, and 16. To maintain consistency, we initialized the tokens randomly. Our results, illustrated in Figure \ref{fig:context_length}, show that APPLeNet outperforms the respective state-of-the-art methods by $0.7\%$, $2.4\%$, and $0.4\%$ for context lengths 1, 4, and 16, respectively. Additionally, we found that APPLeNet achieved the best performance with $\mathcal{M}=4$ among all the context length settings.

\noindent\textbf{Sensitivity to the multi-scale features:}
In this study, we aimed to assess the sensitivity of APPLeNet to visual content features obtained from multiple layers of $f_v$. We utilized two CLIP vision backbones based on ResNet-50 and ViT and increased the number of feature layers in calculating $\hat{F}(x)$. As shown in Figure \ref{fig:class_multi}, incorporating more visual embedding layers to extract content features yielded improved performance, with a monotonically increasing trend. It is worth noting that all experiments included consideration of the style feature $\mu$.

To further highlight the importance of the injection block for intelligent multi-scale feature aggregation, we compare APPLeNet with the multi-scale version of CoCoOp \cite{cocoop}. Specifically, we passed $\hat{F}(x)$ to the meta-network ($\pi$) to devise the Multi-Scale (MS) - CoCoOp (see Table \ref{ms_ablate}). For APPLeNet, we considered three variants where we only pass the multi-scale content features $\hat{F}(x)$, the style features $\mu$, and $F(x)$ to the injection block $\mathcal{B}$. Our results clearly demonstrate that our multi-scale feature aggregation approach outperforms MS-CoCoOp significantly. Additionally, the results highlight the benefits of considering style primitives.

\begin{table}[!ht]
\vspace{-0.25cm}
\centering
\scriptsize{
    \centering
    \caption{\small{Ablation of multi-scale features' sensitivity in CoCoOp and APPLeNet. Harmonic mean (H) of base and new classes are considered for comparison. Best results are shown in \textbf{bold}.}}
    \vspace{-0.2cm}
    \scalebox{0.91}{
    \begin{tabular}{lcccc} 
    \toprule
    \multirow{1}{*}{\textbf{Context Vectors}}&\multicolumn{1}{c}{\textbf{PatternNet}}&\multicolumn{1}{c}{\textbf{RSICD}}
    &\multicolumn{1}{c}{\textbf{RESISC45}}
    &\multicolumn{1}{c}{\textbf{MLRSNet}} \\
     
     
    
    \midrule

    MS-CoCoOp & 75.83 & 72.31 & 69.92 & 62.64 \\
    APPLeNet (with MS) & 77.34 & 73.96 & 72.51 & 65.02 \\
    APPLeNet (with $\mu$)  & 76.04 & 72.19 & 69.53 & 63.95 \\ 
    APPLeNet (with MS \& $\mu$) & \textbf{77.55} & \textbf{74.16} & \textbf{72.73} & \textbf{65.66} \\ \bottomrule
    \end{tabular}}\label{ms_ablate}}
    \vspace{-0.2cm}
\end{table}

\noindent\textbf{Effect of CRP loss ($\mathbf{L_{CRP}}$):}
Table \ref{crploss1} shows the results of ablating $\mathbf{L_{CRP}}$ in Equation \ref{eq_7} over two loss functions, namely CRP loss ($\mathbf{L_{CRP}}$) and cross-entropy loss ($\mathbf{L_{ce}}$). Interestingly, we observed that our APPLeNet model achieved an average improvement of approximately $1-3\%$ across all datasets on the B2N generalization task in the presence of $\mathbf{L_{CRP}}$. This result justifies the significant role of $\mathbf{L_{CRP}}$ in ensuring the distinctiveness in [\texttt{$c'_1$,$\cdots$,$c'_{\mathcal{M}}$}] so that they do not convey redundant information.
\begin{table}[!ht]
\centering
\scriptsize{
    \centering
    \caption{\small{Ablation of APPLeNet with and without CRP ($\mathbf{L_{CRP}}$) loss in Equation \ref{eq_7}. Harmonic mean (H) of base and new classes are considered for comparison. Best results are shown in \textbf{bold}.}}
    \vspace{-0.2cm}
    \scalebox{0.96}{
    \begin{tabular}{lcccc} 
    \toprule
    \multirow{1}{*}{\textbf{APPLeNet}}&\multicolumn{1}{c}{\textbf{PatternNet}}&\multicolumn{1}{c}{\textbf{RSICD}}
    &\multicolumn{1}{c}{\textbf{RESISC45}}
    &\multicolumn{1}{c}{\textbf{MLRSNet}} \\
     
     
    
    \midrule

    without $\mathbf{L_{CRP}}$ & 75.34 & 72.89 & 71.63 & 62.15  \\

    with $\mathbf{L_{CRP}}$ & \textbf{77.55} & \textbf{74.16} & \textbf{72.73} & \textbf{65.66}\\ \bottomrule
    \end{tabular}}\label{crploss1}}
    \vspace{-0.2cm}
\end{table}

\noindent\textbf{Ablation with number of attention modules:}
We conducted an ablation study on the injection block (IB) of our APPLeNet by varying the number of attention modules (AMs) for the single-source multi-target domain generalization task. The results are presented in Table \ref{injection}. We found that APPLeNet with three AMs outperformed the others on the source domain (PatternNetv2 dataset). However, IB with two AMs reported the best performance for the target domains, with at least a numerical improvement of $0.1\%$. It is possible that IB with three AMs suffers from a vanishing gradient problem due to its multiple sigmoid output layers and overfits the data.

\begin{table}[!ht]
\centering
\scriptsize{
    \centering
    \caption{\small{Analysis of the number of attention modules (AMs) used in the injection block for single-source multi-target domain generalization task. We use accuracy metrics as the performance measure. Best results are shown in \textbf{bold}.}}
    \vspace{-0.3cm}
    \scalebox{0.94}{
    \begin{tabular}{ccccc} 
    \toprule
    \multirow{3}{*}{\textbf{No. of AMs}}&\multicolumn{1}{c}{\textbf{Source}}&\multicolumn{3}{c}{\textbf{Target}} \\
     
    \cmidrule(lr){2-2}\cmidrule(lr){3-5}
     
    &\multicolumn{1}{c}{\textbf{PatternNetv2}}&\multicolumn{1}{c}{\textbf{RSICDv2}}&\multicolumn{1}{c}{\textbf{RESISC45v2}}
    &\multicolumn{1}{c}{\textbf{MLRSNetv2}}\\
    
    \midrule

    0 & 93.33 & 76.81 & 80.92 & 72.60 \\

    1 & 95.94 & 79.61 & 81.42 & 73.56  \\

    2 & 96.63 & \textbf{81.03} & \textbf{82.23} & \textbf{74.03} \\
    3 & \textbf{96.77} & 78.85 & 78.21 & 73.92  \\ \bottomrule
    \end{tabular}}\label{injection}}
    \vspace{-0.5cm}
\end{table}

\section{\textbf{Takeaways}}
This paper presents a novel approach, APPLeNet, for prompt learning in CLIP based foundation model for solving three challenging DG tasks in RS. We acknowledge the challenges associated with processing remote sensing scenes, and thus, we propose leveraging the frozen vision backbone of CLIP to generate multi-scale visual content features and batch statistics to generate style properties automatically. We combine visual and learnable text tokens for prompt learning, but since adding visual information can introduce redundancy, we present an anti-correlation regularizer to ensure token distinctiveness.

Our study is the first to extensively evaluate the DG paradigm in remote sensing, and we introduce new benchmarks with comprehensive experimentation. We hope our findings will inspire further research on foundational models for remote sensing applications.

\clearpage
{\small
\bibliographystyle{ieee_fullname}
\bibliography{egbib}
}

\end{document}